# Exploiting Sublanguage and Domain Characteristics in a Bootstrapping Approach to Lexicon and Ontology Creation

**Dietmar Rösner, Manuela Kunze**

Otto-von-Guericke-University Magdeburg,
Institute of Knowledge and Language Processing,
P.O.box 4120,
D-39016 Magdeburg, Germany
{roesner,makunze}@iws.cs.uni-magdeburg.de

**Abstract**

It is very costly to build up lexical resources and domain ontologies. Especially when confronted with a new application domain lexical gaps and a poor coverage of domain concepts are a problem for the successful exploitation of natural language document analysis systems that need and exploit such knowledge sources. In this paper we report about ongoing experiments with 'bootstrapping techniques' for lexicon and ontology creation.

## 1. Introduction

It is very costly to build up lexical resources and domain ontologies. Especially when confronted with a new application domain lexical gaps and a poor coverage of domain concepts are a problem for the successful exploitation of natural language document analysis systems that need and exploit such knowledge sources.

We are confronted with such a situation very often in our work with the XDOC document suite, a collection of tools created to support intelligent processing of corpora of interesting textual documents taken from domains like engineering and medicine. The XDOC document workbench is currently employed in a number of applications. These include:

- knowledge acquisition from technical documentation about casting technology,
- extraction of company profiles from WWW pages,
- analysis of autopsy protocols.

The latter application is part of a joint project with the institute for forensic medicine of our university. The paper is organised as follows: We start with background information about XDOC. Then we sketch characteristics of the sublanguage of autopsy protocols and describe the core idea of our experiments. This is followed by a description of syntactic structures that are currently processed. Then clustering of co-occurrence data and its exploitation is described. A discussion of results and problems and an outlook on future work completes the paper.

## 2. Background: the XDOC document suite

We have designed and implemented the XDOC document suite as a workbench for the flexible processing of electronically available documents in German. The tools in the XDOC document suite(Kunze and Rösner, 2001a), (Kunze and Rösner, 2001b) can be grouped according to their function:

- preprocessing
- structure detection
- POS tagging
- syntactic parsing
- semantic analysis
- tools for the specific application: e.g. information extraction

In the semantic analysis, similar to the POS tagging, the tokens are annotated with their meaning and a classification in semantic categories like e.g. concepts and relations. For the semantic tagging we apply a semantic lexicon. This lexicon contains the semantic interpretation of a word and its case frame combined with the syntactic valence requirements. When we are confronted with a new application domain, the lexical resources must be completed with the domain specific terms. Even semantic resources with broad coverage like the semantic lexicon GermaNet for German (GermaNet-Project-Site, 2002) and Wordnet(Wordnet-Project-Site, 2002) for English, can not cover all terms of all different domains.

### 2.1. Design principles

The work in the XDOC project is guided by the following design principles:

- The tools shall be usable for 'realistic' documents.
  One aspect of 'realistic' documents is that they typically contain domain-specific tokens that are not directly covered by classical lexical categories (like noun, verb, ...). Those tokens are nevertheless often essential for the user of the document (e.g. an enzyme descriptor like EC 4.1.1.17 for a biochemist).

- The tools shall be as robust as possible.
  In general it can not be expected that lexicon information is available for all tokens in a document. This is not only the case for most tokens from 'nonlexical' types – like telephone numbers, enzyme names, material codes, ... –, even for lexical types there will always be 'lexical gaps'. This may either be caused by neologisms or simply by starting to process documents from a new application domain with a new sublanguage. In the latter case lexical items will typically be missing in the lexicon ('lexical gap') and phrasal structures may not or not adequately be covered by the grammar.

- The tools shall be usable independently but shall allow for flexible combination and interoperability.

- The tools shall not only be usable by developers but as well by domain experts without linguistic training.

### 2.2. XML as unifying framework

We have decided to exploit XML (Bray et al., 1998) and its accompanying formalisms (e.g. XSLT (Site, 2002)) and tools (e.g. xt (Clark, 2002) ) as a unifying framework. All modules in the XDOC system expect XML documents as input and deliver their results in XML format.

This decision has positive consequences for many aspects in XDOC. Take e.g. the desideratum that the tools of XDOC shall not only be usable by developers but as well by domain experts without linguistic training. Here XML and XSLT play a major role: XSL stylesheets can be exploited to allow different presentations of internal data and results for different target groups; for end users the internals are in many cases not helpful, whereas developers will need them for debugging.

### 2.3. Bridging lexical gaps

We do not expect extensive lexicon coding at the beginning of an XDOC application. XDOC's POS tagger and syntactic parser have therefore been augmented with a number of techniques for dealing with such 'lexical gaps'.

For POS tagging we exploit the morphology component MORPHIX (Finkler and Neumann, 1988): If a token in a German text can be morphologically analysed with MORPHIX the resulting word class categorisation is used as POS information. Note that this classification need not be unique. Since the tokens are analysed in isolation multiple analyses are often the case. Some examples: the token 'der' may either be a determiner (with a number of different combinations for the features case, number and gender) or a relative pronoun, the token 'liebe' may be either a verb or an adjective (again with different feature combinations not relevant for POS tagging).

MORPHIX's coverage can be characterised as follows: most closed class lexical items of German as well as all irregular verbs are covered. The coverage of open class lexical items is dependent on the amount of coding. The paradigms for e.g. verb conjugation and noun declination are fully covered but to be able to analyze and generate word forms their roots need to be included in the MORPHIX lexicon.

Due to lexical gaps some tokens will not get a MORPHIX analysis, at least at the beginning of an XDOC application. We then employ two techniques: We first try to make use of heuristics that are based on aspects of the tokens that can easily be detected with simple string analysis (e.g. upper-/lowercase, endings, ...) and/or exploitation of the token position relative to sentence boundaries (detected in the structure detection module). If a heuristic yields a classification the resulting POS class is added together with the name of the employed heuristic (marked as feature SRC, cf. example 1). If no heuristics are applicable we classify the token as member of the class unknown (tagged with XXX).

To keep the POS tagger fast and simple the disambiguation between multiple POS classes for a token and the derivation of a possible POS class from context for an unknown token are postponed to syntactic processing (cf. below).

## 3. Bootstrapping in a new domain

XDOCs most recent application is part of a joint project with the institute for forensic medicine of our university. The medical doctors there are interested in tools that help them to exploit their huge collection of several thousand autopsy protocols for their research interests. The confrontation with this corpus from a new domain has stimulated experiments with 'bootstrapping techniques' for lexicon and ontology creation.

### 3.1. The core idea

The core idea is the following:

When you are confronted with a new corpus from a new domain, try to find linguistic structures in the text that are easy to detect automatically and that allow to classify unknown terms in a robust manner both syntactically as well as on the knowledge level. Take the results from a run of these simple but robust heuristics as an initial version of a domain dependent lexicon and ontology. Exploit these initial resources to extend the processing to more complicated linguistic structures in order to detect and classify more terms of interest automatically.

An example: In the sublanguage of autopsy protocols (in German) a very telegrammatic style is dominant. Condensed and compact structures like the following are very frequent:

- Harnblase leer. (Urinary bladder empty.)

- Harnleiter frei. (Ureter free.)

- Nierenoberflaeche glatt. (Surface of kidney smooth.)

- Vorsteherdruese altersentsprechend. (Prostate corresponding to age.)

- ...

These structures can be abstracted syntactically as <Noun><Adjective><Fullstop> and as semantically <Anatomic-entity><Attribute-value><Fullstop>. Furthermore they are easily detectable.

In our experiments we have exploited this characteristic of the corpus extensively to automatically deduce an initial lexicon (with nouns and adjectives) and an initial ontology (with concepts for anatomic regions or organs and their respective features and values).

### 3.2. A sublanguage analysis of autopsy protocols

The telegrammatic style of autopsy protocolls results in a preference for 'verbless' structures. It is e.g. much more likely that a finding like 'the mouth was open' is expressed as 'Mund geoeffnet.' (mouth open) although a more verbose paraphrase like 'Der Mund ist geoeffnet.' may occur sometimes.

Another consequence of the style is a preference for noun compounds in contrast to semantically equivalent noun phrases.

When referring to a concept like 'weight of the liver' the noun compound 'Lebergewicht' is more likely than the noun phrase 'Gewicht der Leber'. This generalizes for the weight of other organs: 'Organgewicht' is more likely than the noun phrase 'Gewicht des/der X'.

The need for contextual interpretation of terms may be seen as another consequence of the style. In local context with an organ as topic generic terms like 'Gewicht' (weight) or 'Durchmesser' (diameter) have to be interpreted as referring to the object in focus, i.e. the organ.

### 3.3. Refinements of the initial approach

In our corpus it is very likely that a syntactic structure of the type <Noun><Adjective><Fullstop> can semantically be interpreted as <Anatomic-entity><Attribute-value><Fullstop>, but there are exceptions. An example: 'Flachschnitt unauffaellig.' Here the noun does not denote an anatomic entity, but is referring to a diagnostic procedure in autopsy. On the other hand the adjective co-occurs with anatomic entities as well.

So the initial approach needs refinement: as long as the number of exceptions of a simple pattern (here: <Noun> <Adjective> <Fullstop>) in a heuristic remains small the exceptions (here: noun 'Flachschnitt') are simply checked first before the heuristic is applied for all cases in which the exceptions are not present.

### 3.4. Exploitation of syntactic constraints

Pattern based analysis is a first step only. For full syntactic parsing we apply a chart parser based on context free grammar rules augmented with feature structures. The output of a robust POS tagger is used as input to parsing. The POS tagger works on token in isolation. Its output may contain:

- multiple POS classes,
- unknown classes of open world tokens and
- tokens with POS class, but without or only partial feature information.

**Example 1** *unknown token classified as noun with heuristics*

```
<NP TYPE="COMPLEX" RULE="NPC3" GEN="FEM"
       NUM="PL" CAS="_">
  <NP TYPE="FULL" RULE="NP1" CAS="_"
         NUM="PL" GEN="FEM">
    <N SRC="UNG">Blutanhaftungen</N>
  </NP>
  <PP CAS="DAT">
    <PRP CAS="DAT">an</PRP>
    <NP TYPE="FULL" RULE="NP2" CAS="DAT"
           NUM="SG" GEN="FEM">
      <DETD>der</DETD>
      <N SRC="UC1">Gekroesewurzel</N>
    </NP>
  </PP>
</NP>
```

The latter case results from some heuristics in POS tagging that allow to assume e.g. the class noun for a token but do not suffice to detect its full paradigm from the token (note that there are approximately two dozen different morphosyntactic paradigms for noun declination in German).

For a given input the parser attempts to find all complete analyses that cover the input. If no such complete analysis is achievable it is attempted to combine maximal partial results into structures covering the whole input (Rösner, 2000).

A successful analysis may be based on an assumption about the word class of an initially unclassified token (tagged XXX). This is indicated in the parsing result (feature AS) and can be exploited for learning such classifications from contextual constraints. In a similar way the successful combination from known feature values from closed class items (e.g. determiners, prepositions) with underspecified features in agreement constraints allows the determination of paradigm information from successfully processed occurrences. In example 2 features of the unknown word "Mundhoehle" (mouth) could be derived from the features of the determiner within the PP (e.g. gender feminine).

**Example 2** *unknown token classified as adjective and features derived through contextual constraints*

```
<NP TYPE="COMPLEX" RULE="NPC3" GEN="MAS" NUM="SG"
     CAS="NOM">
  <NP TYPE="FULL" RULE="NP3" CAS="NOM" NUM="SG"
         GEN="MAS">
    <DETI>kein</DETI>
    <XXX AS="ADJ">ungehoeriger</XXX>
    <N>Inhalt</N>
  </NP>
  <PP CAS="DAT">
    <PRP CAS="DAT">in</PRP>
    <NP TYPE="FULL" RULE="NP2" CAS="DAT" NUM="SG"
           GEN="FEM">
      <DETD>der</DETD>
      <N SRC="UC1">Mundhoehle</N>
    </NP>
  </PP>
</NP>"
```

The grammar used in syntactic parsing is organised in a modular way that allows to add or remove groups of rules. This is exploited when the sublanguage of a domain contains linguistic structures that are unusual or even ungrammatical in standard German.

**Example 3** *Excerpt from syntactic analysis*

```
<PP CAS="AKK">
  <PRP CAS="AKK">auf</PRP>
  <NP TYPE="COMPLEX" RULE="NPC3" GEN="MAS" NUM="SG"
     CAS="AKK">
    <NP TYPE="FULL" RULE="NP1" CAS="AKK" NUM="SG"
       GEN="MAS">
      <N>Flachschnitt</N>
    </NP>
    <PP CAS="AKK">
      <PRP CAS="AKK">in</PRP>
      <NP TYPE="FULL" RULE="NP2" CAS="AKK" NUM="SG"
         GEN="NTR">
        <DETD>das</DETD>
        <N>Gewebe</N>
      </NP>
    </PP>
  </NP>
</PP>
```

## 3.5. Beyond simple patterns

At the time we work with a 'light' grammar of 40 rules. This grammar contains basic rules (for the analysis of noun phrases and preposition phrases) and specific rules, based on the patterns of the sublanguage.

We have just started to extract binary relations from completely parsed sentences. Following patterns of the sublanguage are analysed in this manner: Simple structures like: <NP> <Adjective> <Fullstop> will be analysed as <Anatomic-entity> <Attribute-value> <Fullstop>.

**Example 4** *For example: 'Gehirngaenge frei.'. The Analysis returns:*

```
<RATT-V>
    <ENTITY>Gehoergaenge</ENTITY>
    <VALUE CNT="1">frei</VALUE>
</RATT-V>
```

All results of this analysis are also marked as XML structure. The attribute 'CNT' contains the number of occurences of the attribute value in context with the anatomic entity. A similar pattern is the structure <NP> 'ist|sind'[1] <Adjective>|<Verb> <Fullstop>.

**Example 5** *For example: 'Gangsysteme sind frei.' or 'Augen sind geschlossen'. The Analysis returns:*

```
<RATT-V>
    <ENTITY>Gangsysteme</ENTITY>
    <VALUE CNT="1">frei</VALUE>
</RATT-V>
<RATT-V>
    <ENTITY>Augen</ENTITY>
    <VALUE CNT="1">geschlossen</VALUE>
</RATT-V>
```

Further on we analyse structures which contain more than attribute and domain entity. We extended our analyses to structures, which e.g. contain a modifier like 'sehr' (very) or a negator like 'nicht' (not) and other adjectives.

**Example 6** *Result of the example: 'Brustkorb nicht sehr breit.'*

```
<RATT-V>
    <ENTITY>Brustkorb</ENTITY>
    <VALUE CNT="1">nicht-sehr-breit</VALUE>
</RATT-V>
```

Here the attribute is compounded of a series of words from different wordclasses, because at the time we work with binary relations only. In ongoing work we will further detail this semantic interpretation. In addition we analyse complex structures like coordinated structures. There exist various pattern, e.g. <NP> <Adjective>|<Verb> 'und' <Adjective>|<Verb><Fullstop>. These structures are interpreted as <Anatomic-entity> <Attribute-value1> 'and' <Anatomic-entity> <Attribute-value2><Fullstop>.

**Example 7** *For example: 'Beckengeruest festgefuegt und unversehrt.'. The result is:*

```
<RATT-V>
    <ENTITY>Beckengeruest</ENTITY>
    <VALUE CNT="1">festgefuegt</VALUE>
    <VALUE CNT="1">unversehrt</VALUE>
</RATT-V>
```

---

[1] is|are, | expresses alternatives in pattern

The inverse structure (the coordination at the beginning of the pattern) e.g. <Adjective> 'und' <Adjective> <NP> <Fullstop> can also be analysed.

**Example 8** *For example: 'Akute und chronische Erweiterung des Herzens.'*

```
<RATT-V>
    <ENTITY>Erweiterung des Herzens</ENTITY>
    <VALUE CNT="1">akute</VALUE>
    <VALUE CNT="1">chronische</VALUE>
</RATT-V>
```

Another coordinated pattern is <NP> 'und' <NP> <Adjective>|<Verb> <Fullstop>. The semantic interpretation is similar to the analysis of the simple structures: <Anatomic-entity1> <Attribute-value> 'and' <Anatomic-entity2><Attribute-value><Fullstop>.

**Example 9** *For example: 'Rippen und Wirbelsaeule intakt.' The result is:*

```
<RATT-V>
    <ENTITY>Rippen</ENTITY>
    <VALUE CNT="1">intakt</VALUE>
</RATT-V>
<RATT-V>
    <ENTITY>Wirbelsaeule</ENTITY>
    <VALUE CNT="1">intakt</VALUE>
</RATT-V>
```

The pattern, like the example 'Leber und Niere ohne Besonderheiten.'('Liver and kidney without findings.'), differs from the last described structures in the kind of the attribute. In this structure the attribute is described by a preposition phrase. The analysis returns

**Example 10** *Result of 'Leber und Niere ohne Besonderheiten.':*

```
<RATT-V>
    <ENTITY>Leber</ENTITY>
    <VALUE CNT="1">ohne Besonderheiten</VALUE>
</RATT-V>
<RATT-V>
    <ENTITY>Niere</ENTITY>
    <VALUE CNT="1">ohne Besonderheiten</VALUE>
</RATT-V>
```

## 4. Ontology creation

### 4.1. Analysis of co-occurrence data

Co-occurrence data are used for clustering: We start e.g. with an adjective token that is related to a single noun type only in the analysed data.

If - again within the corpus given - this noun co-occurs only with this very adjective then the relation between the noun's concept and the property denoted by the adjective is very strong. It may even be the case that the adjective-noun-combination is a name like fixed phrase.

If the noun co-occurs with other adjectives as well it is interesting to uncover the relation between the adjectives and denoted properties respectively.

There are a number of possibilities:

- Two adjectives may be used as 'quasi-synonyms',
- Adjectives may be in an antinomy relation,
- Adjectives may refer to discrete values of a property that are linearly ordered on a scale,

- Adjectives refer to values of different properties.

We can proceed in a zig-zag-manner:

We have started with a single adjective and checked for its co-occurring noun. We then asked for other adjectives co-occurring with this noun. In the next step we extend the set of nouns with those nouns that co-occur with at least one of the adjectives in the adjective set.

Then we can extend the adjective set accordingly. The process will definitely stop if in a step the set to be expanded (either the noun or the adjective set) is no longer growing and has thus reached a fixed point.

As soon as the zig-zag-procedure adds an adjective to the adjective set thats co-occurs with many nouns of different type then in the next step, when the co-occurring adjectives of all these nouns are added, we may produce (nearly) a full covering of all adjectives and of all nouns respectively.

### 4.2. Exploiting co-occurrence information

#### 4.2.1. Concept detection

A noun phrase of the type <Adj> <Noun> may be like the name of a concept but this does not always hold and depends on usage.

An example: 'fluessige Galle' as in 'In der Gallenblase fluessige Galle' is a property value, not a name. On the other hand 'harte Hirnhaut' is to be treated as nameing a concept. This can be inferred from the usage of the NP 'harte Hirnhaut' in structures of the type

<NP><Adj> like 'harte Hirnhaut perlmuttergrau'.

#### 4.2.2. Concept classification

Currently linguistic structures are mapped into binary relations. An example:

Harte Hirnhaut grauweiss.

is an application of the grammar rule with

<NP> <Adjective> <Fullstop>

as right hand side. This establishes a <Property> <Concept> pair.

If we invert this relation (i.e. give a listing of all property values that co-occur – with number of occurrences above a threshold – with the concept) this yields:

Harte Hirnhaut: glaenzend, grauweiss, perlmuttergrau, weisslich-gelblich-verfaerbt, intakt, grauroetlich, blaeulich-durchscheinend

If we analyze these adjectives (and compounded adjective groups) we find the following:

- there is one very generic property 'intakt' (engl. 'intact') that is usable with almost any anatomic-entity

- the adjective 'glaenzend' is characterising the visual appearance of the brain skin as shiny

- all other adjectives denote a variety colors

Thus the brain skin can be classified as an anatomic-entity whose color values are relevant in autopsy reports.

#### 4.2.3. Concept grouping

Clustering of co-occurrence data allows to detect candidates for semantic groups as well as synonyms and/or paraphrases.

- 'spiegelnd': 'Herzueberzug', 'Lungenueberzug'

- 'unversehrt': 'Haut des Rueckens', 'Stirnhaut'

- 'frei': 'Gehoergange', 'Ausfuehrungsgang', 'Kehlkopfeingang'

All concepts co-occurring with 'frei' are of the type tube.

### 4.3. Ontological relations

What ontological relations can be inferred?

- Is-a: Leber Is-a Organ

- Part-of: Schleimhaut Part-of Magen (generalized Schleimhaut Part-of Organ)

- other n-nary relations: e.g. 'nicht widernatuerlich beweglich'

Further on we can find a classification of relations resp. the domain range of an relation. For example the relation 'geoeffnet' (opened) can be changed by modifier in the attribute-value

- 'geoeffnet'

- 'spaltweit-geoeffnet'

- 'spaltfoermig-geoeffnet'

- 'geschlossen' (as opposite to 'geoeffnet')

## 5. Discussion

Our current work is of an investigative nature. The size of the corpus is still small. It is planned to apply the techniques developed with the initial corpus to the collection of several thousand protocols. The number of occurrences is still small and statistical methods are therefore not yet adequate. Even if quantitative measures are not applicable on the basis of this corpus occurrence data can be interpreted qualitatively.

Since we have just recently started with the domain of autopsy protocols there are e.g. still gaps in grammar coverage and in the tagging process (not every unknown word can be classified by heuristics). In the corpus currently approx 37 % of the sentences and telegrammatic structures can be fully processed (i.e. get at least one reading covering the structure as a whole; multiple readings are possible.) Experiments with the full corpus will allow to evaluate how reliable the results are.

The telegrammatic style results in shorter and – on the first sight – 'simpler' linguistic structures. As a trade-off these structures are less constrained and this e.g. complicates the derivation of morphosyntactic features from context or makes inferred results less reliable.

An example: If 'Nieren' is an unknown token the full sentence 'Die Nieren sind unversehrt' allows to infer that the token is a plural form, the same inference is not possible from the telegrammatic version 'Nieren unversehrt'.

### 5.1. XDOC as a workbench

We are aiming at a workbench with a rich functionality but we do not expect a fully automatic and autonomous solution. The user shall be supported as good as possible but s/he will still be involved in the process.

Our approach is interactive. The user has to confirm suggestions from the system. He is accepting or refusing, but can delegate searching, comparing, counting etc. to the system.

### 5.2. Acquisition of domain knowledge

Some findings in autopsy protocols are results of measurements: values of weights, sizes, diameters etc. are reported.

This allows to collect 'typical values' and to gain distributions for ranges of values.

For weights a typical pattern is:

<organ>gewicht <number> g.

'Lebergewicht ... g'

From the texts we derived the range of the weight-relation for example for the organ kidney as 135 g to 270 g (in a medical manual the weight of the kidney is defined in the range of 120 g to 300 g).

Sometimes contextual interpretation is necessary:

<organ><property-value>. Gewicht <number> g.

Here the generic term 'Gewicht' (weight) has to be interpreted as referring to the organ in focus.

Similar constructions are employed for other indicators like diameters.

### 5.3. Future work:

For the quality of inferences the detection of synonyms and paraphrases plays a major role, e.g. 'Blase' and 'Harnblase' do refer to the same organ, 'Stirnhaut' and 'Haut der Stirn' denote the same region: the skin of the forehead.

A general solution for coordinated structures will be necessary.

A subtype of coordinated structures includes truncation of compounds. An example: 'Wangen- und Kinnpartie unauffaellig.' The reconstruction of the untruncated term is not always as simple as in the example. For this task we need an approach similar to the one described in (Buitelaar and Scaleanu, 2002). It must not only be analysed which is the semantic meaning of the word, but rather which is the word, which was truncated. One criterion is, that the words must have the same semantic category.

A general component for the semantic treatment of noun compounds is needed. This will have to interact with contextual interpretation. In an example like

*24. Hirngewicht 1490 g. Windungen abgeflacht, Furchen verstrichen. ...*

it has to be detected that with the reference to the weight of the brain ('Hirngewicht') the brain is established as topic and that the terms 'Windungen' and 'Furchen' are referring to findings about the brain's visible appearance.

Autopsy protocols are written in a way such that the course of the autopsy is directly reflected in discourse structure. The autopsy on the other hand follows anatomic structures and their neighbourhood relations. In local contexts we both find part-of relations between anatomic structures as well as neighbourhood relations.

The analysis of noun phrases needs to be more fine grained. Structures like 'Haut des Rueckens' or 'Haut ueber der Nase' should e.g. be interpreted as localisation information that is specifying regions of the skin (here: 'skin of the back' and 'skin of the nose').